# Data-driven evolutionary algorithm for oil reservoir well-placement and control optimization

Guodong Chen[a], Xin Luo[a], Jimmy Jiu Jiao[a,*], Xiaoming Xue[b]

a Department of Earth Sciences, The University of Hong Kong, Hong Kong, China

b Department of Computer Science, City University of Hong Kong, Hong Kong, China

Corresponding author: Jimmy Jiu Jiao (jjiao@hku.hk)

## Abstract

Optimal well placement and well injection-production are crucial for the reservoir development to maximize the financial profits during the project lifetime. Meta-heuristic algorithms have showed good performance in solving complex, nonlinear and non-continuous optimization problems. However, a large number of numerical simulation runs are involved during the optimization process. In this work, a novel and efficient data-driven evolutionary algorithm, called generalized data-driven differential evolutionary algorithm (GDDE), is proposed to reduce the number of simulation runs on well-placement and control optimization problems. Probabilistic neural network (PNN) is adopted as the classifier to select informative and promising candidates, and the most uncertain candidate based on Euclidean distance is prescreened and evaluated with a numerical simulator. Subsequently, local surrogate model is built by radial basis function (RBF) and the optimum of the surrogate, found by optimizer, is evaluated by the numerical simulator to accelerate the convergence. It is worth noting that the shape factors of RBF model and PNN are optimized via solving hyper-parameter sub-expensive optimization problem. The results show the optimization algorithm proposed in this study is very promising for a well-placement optimization problem of two-dimensional reservoir and joint optimization of Egg model.

# 1 Introduction

Obtaining the optimal oil reservoir development plan, such as determining the locations of wells and the flow rates or bottom-hole-pressure of wells, plays an important role in the field development processes [1-3]. Well-placement and well injection-production optimization can maximize the financial profits during the oilfield development period by improving the sweep efficiency of the water-flooding reservoir. Optimal well location and well-control scheme can be impacted by many factors, such as the heterogeneous permeability of the reservoir, the properties of fluids and the developing history of the reservoir [4, 5]. Well-placement optimization is a highly nonlinear problem and the decision variables are integers [6]. Such optimization problems are typically computationally expensive, as the objective function has to be evaluated with time consuming numerical simulation in the optimization processes [7, 8].

Gradient-based algorithms have been developed to solve well placement optimization problems efficiently [9-11]. Handels et al. [12] adopted adjoint-based gradient algorithm on constrained well-placement optimization problems. Forouzanfar and Reynolds [13] used a gradient-based algorithm to solve the joint optimization of well number and locations and controls. However, these algorithms are easy to get stuck into local optima and the gradient calculated by adjoint or finite-difference method may not be precise for such discrete problems. Derivative-free algorithms, such as genetic algorithm (GA) [14-17], particle swarm optimization (PSO) [18-20], differential evolution (DE) [21, 22] and covariance matrix adaptation-evolution strategy (CMA-ES) [2, 23], have been commonly used for well-placement optimization problems, as these heuristic methods are able to jump out from the local optimum and converge to global optimum [24-26]. Beckner and Song [24] adopted simulated annealing method to determine optimal economic well control and placement. Burak et al. [14] presented hybrid GA to optimize the nonconventional well type, location, and trajectory. Onwunalu and Durlofsky [19] adopted PSO for well placement and type. Ding et al. [18] proposed a hybrid objective function with PSO as optimizer for well-placement problems. Awotunde et al. [2] used CMA-ES as the optimizer to determine the well location and considered minimum well spacing as constraint. Nevertheless, these algorithms need to consume a large number of simulation runs, especially for high-dimensional problems, before locating the global optimum which is computationally prohibitive [21].

Surrogate model (also called proxy) has gained increasing attention recently due to the promising ability on reducing the number of simulation runs during the optimization processes [27-29]. Various surrogate methods have been developed to save computational time, e.g., reduced order model (ROM) [30], capacitance-resistance model (CRM) [31] and machine learning methods [32-34]. Jansen and Durlofsky [30] developed proper orthogonal decomposition to reduce the order of model and accelerate the computation of single simulation. However, ROM needs to obtain the explicit information of the simulator which is impossible for commercial simulator like Eclipse. Yousef et al. [31] presented capacitance-resistance model to quantify connectivity and degree of fluid storage between wells. Zhao et al. proposed a physical-based interwell-numerical-simulation model (INSIM) using mass-material-balance and front-tracking equations for the control units [32]. However, INSIM and CRM are less faithful and the accuracy is poor especially on well-placement optimization problems.

Machine learning methods, which are computationally cheap mathematical models, can approximate the input/output relationship between the decision variables and the objective function [27, 35]. Commonly used machine learning methods as surrogate are: Gaussian process (GP) [36, 37], radial basis function (RBF) [38], polynomial regression surface (PRS) and support vector machine (SVM) [39, 40]. Taking the prediction and the uncertainty of sample points into consideration, kriging-based infill criteria, such as lower confidence bound (LCB) [37], expected improvement (EI) and probability of improvement (PoI), can balance the exploration and the exploitation. Bernardo et al. [41] used GP as surrogate and sequential quadratic programming as optimizer to solve water-flooding well-control optimization problems. Redouane et al. [28] presented GP and GA combined with hybrid constraint-handling strategy on well-placement problems. Guo and Reynolds [42] proposed a workflow with support vector regression as surrogate and stochastic simplex approximate gradient as optimizer to estimate the optimal well-control scheme for the robust production optimization problems. Nevertheless, when dealing with high-dimensional production optimization or joint optimization problems, the performance of surrogate-assisted evolutionary algorithm deteriorates drastically.

High-dimensional expensive optimization is a challenging problem and many researchers have made efforts to solve this problem recently. Liu et al. [43] proposed surrogate-assisted differential evolution (GPEME) using GP combined with dimensionality reduction method on solving 20-50 dimensional problems. Sun et al. [44] combined a surrogate-assisted PSO algorithm and a surrogate-assisted social learning PSO algorithm (SACOSO) to cooperatively guide the search. Yu et al. [45] presented surrogate-assisted hierarchical particle swarm optimization (SHPSO) algorithm for 30-100 dimensional benchmark functions. Wang et al. [46] developed an evolutionary sampling assisted optimization (ESAO) with RBF as surrogate and DE as optimizer on solving 20-100 dimensional benchmark functions. Cai et al. [47] proposed a generalized surrogate-assisted genetic algorithm (GSGA), consisting of surrogate-based local search, surrogate-guided GA updating mechanism and a kriging-based prescreening strategy. Li et al. [48] used a boosting strategy for model management and a localized data generation method to alleviate small sample problem on solving 10-100 dimensional problems (BDDEA-LDG). Chen et al. [27] presented a hierarchical surrogate-assisted evolutionary algorithm (EHSDE) combining global prescreening strategy and local acceleration strategy on solving 20-100 dimensional problems, and the performance is promising in comparison with other state-of-the-art algorithms. Classification-based machine learning method has also been introduced as surrogate. Wei et al. [49] proposed a classifier-assisted level-based learning swarm optimizer (CA-LLSO) using the gradient boosting gradient classifier and the level-based learning swarm optimizer for 20-300 dimensional benchmark functions. However, when dealing with high-dimensional expensive problems, these methods still suffer from the curse of dimensionality.

The objective of this study is to use machine learning methods guiding the search of evolutionary algorithm to accelerate the convergence and reduce the number of simulation runs on solving well-placement optimization and joint optimization problems. A novel and efficient data-driven evolutionary algorithm is developed to alleviate the curse of dimensionality and accelerate the convergence during the optimization process. The proposed algorithm is called generalized data-driven differential evolutionary algorithm (GDDE) and consists of two stages. In the first stage, probabilistic neural network (PNN) is adopted as classifier to select informative and promising candidates, and the most uncertain candidate based on Euclidean distance evaluated with simulator. In the second stage, surrogate with RBF is built in a small promising area and the optimum of the surrogate is evaluated by numerical simulator to accelerate the convergence. The shape factor of RBF model is optimized via solving a hyper-parameter optimization problem. To illustrate the performance, the proposed algorithm is tested on a two-dimensional (2D) reservoir model and Egg model.

The main contributions of this work are: 1) The study adopts both classification model and interpolation model as the surrogates to guide evolutionary algorithm in solving well-placement and joint optimization problems. Concretely, a classifier is adopted to identify the promising individuals from the offsprings. The most uncertain individual is selected from the promising individuals to evaluate with real function evaluation.

Besides, RBF is adopted to approximate the landscape of the objective function at the local promising area to accelerate the convergence. 2) To obtain the optimal shape factor of RBF, PRS is adopted as surrogate to minimize the leave-one-out crossover validation error. After getting the optimal shape factor, the accuracy of the local surrogate can be improved greatly and the convergence speed can be accelerated significantly.

The rest of this paper is organized as follows. Firstly, the well-placement optimization and joint optimization problems are introduced in section 2. Then the related techniques involved in the proposed algorithm are introduced in section 3. Subsequently, the framework of the proposed algorithm is described in section 4 and the experimental results are discussed in section 5. Finally, the conclusion is provided.

## 2 Well-placement and control optimization problem

In recent years, joint optimization of well-placement and control problems have gained increasing interest to identify the optimal well locations and corresponding control scheme [50, 51]. Traditional methods on such problems use a sequential way to determine the optimal development scheme which is easy to suffer from local optimum [52, 53]. Li and Jafarpour [54] proposed to use derivative-free random stochastic search method for well-placement and gradient-based algorithm for well control problem alternatively.

In this study, a joint rather than a sequential manner is adopted to achieve better water-flooding reservoir development plan and maximize the profit during the development period by improving the sweep efficiency of the reservoir [55, 56].

The joint optimization of well placement and control problem can be defined as follows:

$$\max_{\mathbf{x}_1, \mathbf{x}_2} NPV(\mathbf{x}_1, \mathbf{x}_2) \tag{1}$$

subject to

$$\mathbf{x}_{1l} \leq \mathbf{x}_1 \leq \mathbf{x}_{1u}, \mathbf{x}_1 \in \mathbf{Z}^{n_1} \tag{2}$$

$$\mathbf{x}_{2l} \leq \mathbf{x}_2 \leq \mathbf{x}_{2u}, \mathbf{x}_2 \in \mathbf{R}^{n_2} \tag{3}$$

where $NPV(\mathbf{x}_1, \mathbf{x}_2)$ denotes the objective function of the problem; $\mathbf{x}_1$ denotes the discrete integer vector of well locations to be optimized; $n_1$ is the number of variables of well locations; $\mathbf{x}_2$ denotes the continuous vector of well control (i.e., the flow rate or the bottom hole pressure of each well); $n_2 = n_t \times (n_p + n_i)$ is the number of variables of well control (i.e., the number of time steps multiply the number of wells); $\mathbf{x}_{1l}$ and $\mathbf{x}_{1u}$ are the lower and upper boundaries of well locations, respectively; $\mathbf{x}_{2l}$ and $\mathbf{x}_{2u}$ are the lower and upper boundaries of well control variables, respectively. Note that all the wells to be optimized are vertical and only bound constraints are considered in this work. $\mathbf{x}_1$ is predefined for well-placement problems, and $\mathbf{x}_2$ is predefined for well control problems. The simplified objective NPV is defined as follows:

$$NPV(\mathbf{x}_1, \mathbf{x}_2) = \sum_{t=1}^{k} \frac{[Q_{o,t} \cdot r_o - Q_{w,t} \cdot r_w - Q_{i,t} \cdot r_i] \cdot \Delta t}{(1+b)^{p_t}} \tag{4}$$

where $k$ denotes the number of time steps; $\Delta t$ is the length of each time step; $Q_{o,t}$, $Q_{w,t}$ and $Q_{i,t}$ are oil production rate, water production rate and water injection rate at $t^{th}$ time step, respectively; $r_o$ is the oil price while $r_w$ and $r_i$ are the costs of water removal and water injection, respectively; $b$ is the discount rate; $p_t$ is the elapsed time in years.

# 3 Related techniques

## 3.1 Probabilistic neural network

Probabilistic neural network (PNN) is a kind of feed forward neural network consisting of four layers, i.e., input layer, pattern layer, summation layer and output layer [57]. The structure of PNN is shown in Fig. 1.

The first layer is input layer, used to receive the value from sample points and pass to the network. The number of the neurons of input layer equals the dimension of the optimization problem. The second hidden layer uses RBF as kernel function with training sample points as the neuron center to calculate the distance between input vector and neuron center. The input/output relationship from vector $x$ to the $j^{th}$ neuron of the $i^{th}$ mode is defined as follows:

$$\phi_{ij}(x) = \frac{1}{(2\pi)^{\frac{d}{2}} \sigma^d} \exp[-\frac{(x-x_{ij})^T(x-x_{ij})}{2\sigma^2}] \tag{5}$$

where $d$ is the dimension of the input vector, $\sigma$ is the shape factor and $x_{ij}$ is the neuron vector.

Summation layer can estimate the probability density of each mode with the help of the kernel function. The maximum likelihood of **x** being classified into $C_i$ can be calculated by averaging the value of neuron that is in the same class. The value of the $i^{th}$ neuron of the summation layer can be calculated as follows:

$$p_i(x) = \frac{1}{N_i (2\pi)^{\frac{d}{2}} \sigma^d} \sum_{j=1}^{N_i} \exp[-\frac{(x-x_{ij})^T(x-x_{ij})}{2\sigma^2}] \tag{6}$$

where $N_i$ is the number of training sample points belonging to class $C_i$. After calculating the posteriori probability density, the class with the maximum one will be selected as the output. The output layer uses competitive neurons as classifier.

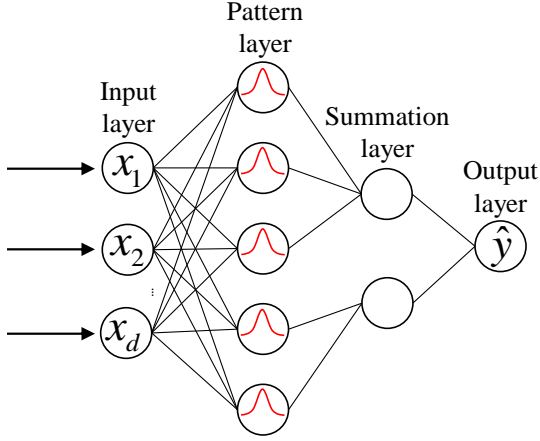 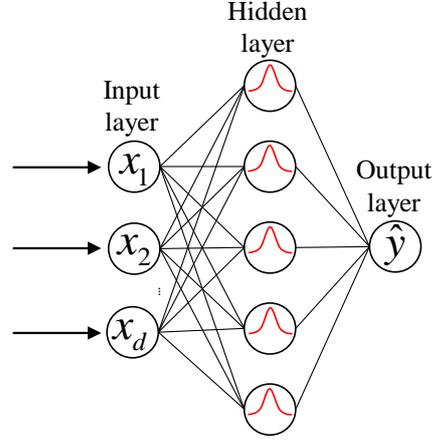

Fig. 1 The structure of PNN    Fig. 2 The structure of RBF

## 3.2 Radial basis function

RBF has been widely adopted as surrogate on solving high-dimensional expensive optimization problems, because it is relatively not sensitive to the dimension and the accuracy is relatively good. RBF is a weighted sum of kernel functions to approximate the landscape of objective function [27]. Suppose $\{(\mathbf{x_i}, f(\mathbf{x_i})), i=1,2,...,n\}$ is the training sample points, $\mathbf{x_i} \in R^d$ and $f(\mathbf{x_i}) \in R$ are the decision vectors and objective function values, respectively. The RBF model is expressed as follows:

$$\hat{f}(\mathbf{x}) = \sum_{i=1}^{n} \omega_i \varphi(\|\mathbf{x} - \mathbf{x_i}\|) \qquad (7)$$

where $\hat{f}$ is the mathematical model of RBF, $\omega_i$ is the $i^{th}$ weight parameter and $\varphi$ is the kernel function. In this work, Gaussian kernel is adopted as basis function:

$$\varphi(x) = \exp(-\frac{x^2}{\sigma^2}) \qquad (8)$$

where $\sigma$ is the shape parameter. The weight parameters can be calculated as follows:

$$\omega = \varphi^{-1} \mathbf{f} \tag{9}$$

**3.3 Polynomial regression surface**

PRS can approximate the objective function $y = f(\mathbf{x})$ in the following form (second order model is adopted):

$$\hat{f}(\mathbf{x}) = \beta_0 + \sum_{i=1}^{d} \beta_i x_i + \sum_{i=1}^{d}\sum_{i=1}^{d} \beta_{ij} x_i x_j \tag{10}$$

where $\boldsymbol{\beta} = [\beta_0, \beta_1, ..., \beta_d, \beta_{11}, ..., \beta_{dd}]^\mathrm{T}$ is the coefficient vector which can be calculated as follows:

$$\boldsymbol{\beta} = (\mathbf{X}^T \mathbf{X})^{-1} \mathbf{X}^T \mathbf{f} \tag{11}$$

$\mathbf{X} = [1 \quad x_1 \quad x_1^2]$ when the dimension equals 1.

**3.4 Differential evolution**

DE, a population-based heuristic algorithm, has been widely used in many complex engineering optimization problems. In this work, DE is adopted as the optimizer and operator to generate offsprings for prescreening. DE consists of initialization, mutation, crossover and selection. Firstly, the population is generated randomly $P = [\mathbf{x_1}, \mathbf{x_2}, ..., \mathbf{x_{NP}}]$ and each individual is a vector with dimension $d$, e.g., the $i^{th}$ individual is $\mathbf{x_i} = (x_{i1}, x_{i2}, ..., x_{id}) \in R^d$. Then mutation operation is conducted with different strategies to generate the donor vector:

DE/best/1

$$\mathbf{v_i} = \mathbf{x_{best}} + Mu(\mathbf{x_{i_1}} - \mathbf{x_{i_2}}) \tag{12}$$

DE/current-to-best/1

$$\mathbf{v_i} = \mathbf{x_i} + Mu(\mathbf{x_{best}} - \mathbf{x_i}) + Mu(\mathbf{x_{i_1}} - \mathbf{x_{i_2}}) \tag{13}$$

where $\mathbf{v_i}$ is the $i^{th}$ donor vector, $\mathbf{x_i}$ denotes the $i^{th}$ individual, $Mu$ is the mutation operator, and $i_1, i_2 \in [1, NP]$ are randomly generated and mutually different integers. In this research, DE/best/1 is adopted as mutation strategy. After mutation, crossover operation is conducted as follows:

$$\mathbf{u}_i^j = \begin{cases} \mathbf{v}_i^j, & if\ (rand \leq CR | j = j_{rand}) \\ \mathbf{x}_i^j, & otherwise \end{cases} \tag{14}$$

where $\mathbf{u}_i^j$ denotes the $j^{th}$ dimension of the $i^{th}$ trial vector, CR denotes the crossover operator, $rand$ is a random number between 0 and 1, and $j_{rand}$ is a random integer from 1 to $d$.

## 4 Methodology

In this study, a novel data-driven evolutionary algorithm called GDDE is proposed for well-placement and joint optimization problems. The main framework of GDDE is introduced first, followed by the details of GDDE.

### 4.1 Main framework of GDDE

The generic diagram of the proposed GDDE is shown in Fig. 3, and the pseudo-code of GDDE is shown in Algorithm 1. Firstly, Latin hypercube sampling (LHS) is used to generate initial sample points. Numerical simulation is used to conduct real function evaluation, and the sample points are added into database D. GDDE consists of two stages. In the first stage, PNN is adopted as classifier to select informative and promising candidates, and the most uncertain candidate based on Euclidean distance is evaluated with simulator. In the second stage, surrogate with RBF is built in a small promising area and the optimum of the surrogate is evaluated by numerical simulator to accelerate the convergence. The shape factor of RBF and PNN models are optimized via solving a hyper-parameter sub-optimization problem. The high-uncertainty and promising individual can be prescreened by Euclidean distance and PNN to explore the search space, while the optimum of the local surrogate can exploit the most promising area to accelerate the convergence. After combining two strategies, the proposed algorithm can balance the exploration and the exploitation and enhance the search ability during the optimization process.

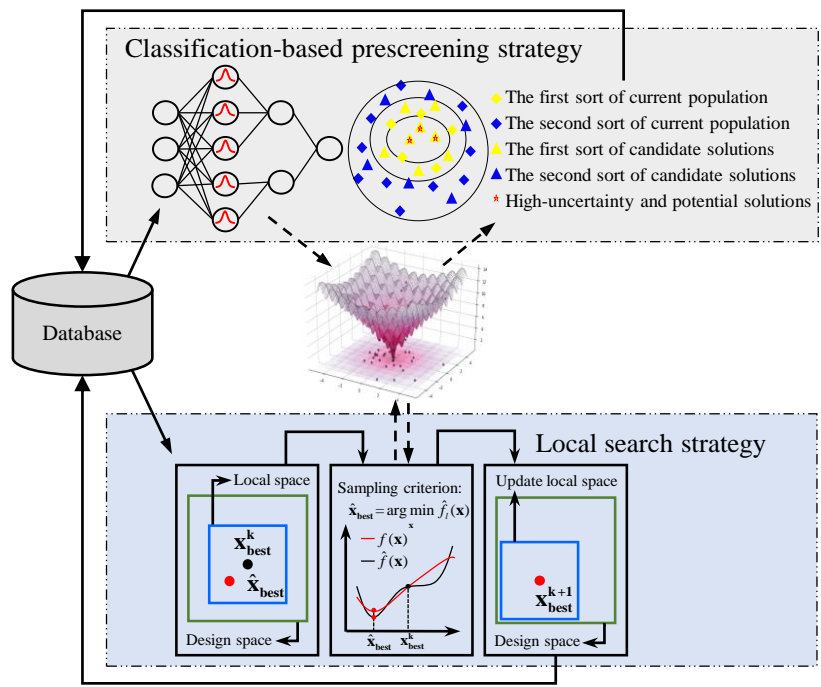

Fig. 3 The framework of the proposed algorithm

**Algorithm 1** Pseudo code of GDDE

01: **Initialization:** Generate $\tau$ initial samples using Latin hypercube sampling from the decision space, conduct function evaluation with numerical simulation, and add into database D.

02: **While** stopping criterion is not met

    // The classification-based prescreening part

03:    Select *NP* best sample points $\mathbf{P}=\{\mathbf{x}_1,\mathbf{x}_2,...,\mathbf{x}_{NP}\}$ from database D;

04:    Sort the population P into two classes: $\lambda$ individuals as the first class $\{\mathbf{x}_k\}^1$, $NP-\lambda$ individuals as the second class $\{\mathbf{x}_k\}^2$;

05:    Train the PNN as classifier with sorted population $\{\mathbf{x}_k\}^1$ and $\{\mathbf{x}_k\}^2$;

06:    Obtain the optimal shape factor $\sigma$ of PNN based on Algorithm 4 (see section 4.4);

07:    Use DE operation to generate *NP* new solutions $\mathbf{u}_1,\mathbf{u}_2,...,\mathbf{u}_{NP}$;

08:    Prescreen the promising sample points belonging to the first class $\{\mathbf{u}_k\}^1$;

09:    Prescreen the most uncertain individual $\mathbf{u}_{unc}$ from the promising individuals $\{\mathbf{u}_k\}^1$ with Euclidean distance;

10:    Evaluate the fitness value of $\mathbf{u}_{unc}$ with simulator;

11:    Record the evaluation of $\mathbf{u}_{unc}$ into database D;

    // The local search part

12:    Choose $\tau$ best sample points $\mathbf{x}_1,\mathbf{x}_2,...,\mathbf{x}_\tau$ as training sample points from database D;

13:    Construct a local RBF model $\hat{f}_l$ with $\tau$ selected sample points, calculate the range of the local space $[\mathbf{lb}_l,\mathbf{ub}_l]$;

14:   Obtain the optimal shape factor $\sigma$ of RBF based on Algorithm 4 (see section 4.4);

15:   Locate the optimum of $\hat{f}_l$ with DE optimizer in the local space;

16:   Evaluate the fitness value of $\hat{\mathbf{x}}_{best}$ with simulator;

17:   Record the evaluation into database D;

18: **End While**

19: **Output**

In the following, the details of GDDE, including classification-based prescreening strategy, local search strategy and sub-optimization of shape parameter will be introduced.

**4.2 Classification-based prescreening strategy**

In this stage, PNN is adopted as classifier to select informative and promising candidates, and the most uncertain candidate based on Euclidean distance is evaluated with simulator. The pseudo-code of this stage is shown in Algorithm 2. Firstly, select $NP$ best sample points $\mathbf{P}=\{\mathbf{x}_1,\mathbf{x}_2,...,\mathbf{x}_{NP}\}$ from the database D as the population. To search for promising individuals, the population P is classified into two classes: $\lambda$ individuals as the first class $\{\mathbf{x}_k\}^1$, $NP-\lambda$ individuals as the second class $\{\mathbf{x}_k\}^2$, and PNN is trained as classifier with sorted population. Then use DE operation to generate $NP$ new solutions $\mathbf{u}_1,\mathbf{u}_2,...,\mathbf{u}_{NP}$. Afterward, a classifier is used to prescreen the promising sample points belonging to the first class $\{\mathbf{u}_k\}^1$, and Euclidean distance is used to prescreen the most uncertain individual $\mathbf{u}_{unc}$ from the promising individuals $\{\mathbf{u}_k\}^1$:

$$g(\mathbf{u}_k) = \min_{\mathbf{u}_k \in \{\mathbf{u}_k\}^1, \mathbf{x} \in \mathbf{D}}(dis(\mathbf{u}_k, \mathbf{x})) \tag{15}$$

where $dis(\mathbf{u}_k, \mathbf{x})$ is the Euclidean distance between trial vector $\mathbf{u}_k$ and training solutions $\mathbf{x}$. The most uncertain solution is selected as follows:

$$\mathbf{u}_{unc} = \arg\max_{\mathbf{u}_k \in \{\mathbf{u}_k\}^1} g(\mathbf{u}_k) \tag{16}$$

where $\mathbf{u}_{unc}$ denotes the most uncertain individual from promising solutions $\{\mathbf{u}_k\}^1$. Subsequently, evaluate the fitness value of $\mathbf{u}_{unc}$ with numerical simulator and record the fitness value of $\mathbf{u}_{unc}$ into database D.

---

**Algorithm 2** Pseudo code of classification-based prescreening strategy

**Input**: Database D, the population size NP, the size of the first class $\lambda$, mutation operator Mu, crossover operator CR.

01: Select *NP* best sample points $\mathbf{P} = \{\mathbf{x}_1, \mathbf{x}_2, ..., \mathbf{x}_{NP}\}$ from database D;

02: Sort the population P into two classes: $\lambda$ individuals as the first class $\{\mathbf{x}_k\}^1$, $NP - \lambda$ individuals as the second class $\{\mathbf{x}_k\}^2$;

03: Train the PNN as classifier with sorted population $\{\mathbf{x}_k\}^1$ and $\{\mathbf{x}_k\}^2$;

04: Use DE operation to generate $NP$ new solutions $\mathbf{u}_1, \mathbf{u}_2, ..., \mathbf{u}_{NP}$;

05: Prescreen the promising sample points belonging to the first class $\{\mathbf{u}_k\}^1$;

06: Prescreen the most uncertain individual $\mathbf{u}_{unc}$ from the promising individuals $\{\mathbf{u}_k\}^1$ with Euclidean distance;

07: Evaluate the fitness value of $\mathbf{u}_{unc}$ with simulator;

08: Record the evaluation of $\mathbf{u}_{unc}$ into database D;

**Output**: Database D.

---

**4.3 Local search strategy**

In this stage, RBF is built as surrogate in a small promising area and the optimum of the surrogate is supplied by DE optimizer and evaluated by numerical simulator to accelerate the convergence. The shape factor of RBF model is optimized via solving a hyper-parameter optimization problem to get a high-quality surrogate. The pseudo-code of this stage is provided in Algorithm 3. Firstly, $\tau$ best sample points $\mathbf{x}_1, \mathbf{x}_2, ..., \mathbf{x}_\tau$ are selected as training sample points from database D. Then a local RBF model is constructed to approximate the landscape of the small promising area. The range of the $i^{th}$ dimension of local surrogate can be calculated as follows:

$$\begin{cases} \mathbf{lb}_l^i = \min(\mathbf{x}_1^i, \mathbf{x}_2^i ..., \mathbf{x}_\tau^i) \\ \mathbf{ub}_l^i = \max(\mathbf{x}_1^i, \mathbf{x}_2^i ..., \mathbf{x}_\tau^i) \end{cases} \quad (17)$$

where $\mathbf{lb}_l^i$ and $\mathbf{ub}_l^i$ is the lower and upper boundaries of the $i^{th}$ variable of the local surrogate, $x_\tau^i$ is the $i^{th}$ variable of the $\tau^{th}$ solution. After determining the local range, the surrogate $\hat{f}_l(\mathbf{x})$ is built in the local space. To accelerate the convergence during the optimization process, the optimum of the local surrogate $\hat{\mathbf{x}}_{best}$ is supplied by DE optimizer according to the following criterion:

$$\hat{\mathbf{x}}_{best} = \arg\min_{\mathbf{x}} \hat{f}_l(\mathbf{x}) \quad (18)$$

Then evaluate the fitness value of the pseudo-optimal solution $\hat{\mathbf{x}}_{best}$ with numerical simulator and record the fitness value of $\hat{\mathbf{x}}_{best}$ into database D. It is worth noting that the promising local space will gradually reduce to the area nearby the optimum until the stop criterion is satisfied.

**Algorithm 3** Pseudo code of local search strategy

**Input**: Database D, size of training sample points $\tau$.

01: Choose $\tau$ best sample points $\mathbf{x}_1, \mathbf{x}_2, ..., \mathbf{x}_\tau$ as training sample points from database D;

02: Construct a local RBF model $\hat{f}_l$ with $\tau$ selected sample points, calculate the range of the local space $[\mathbf{lb}_l, \mathbf{ub}_l]$;

03: Locate the optimum of $\hat{f}_l$ with DE optimizer in the local space;

04: Evaluate the fitness value of $\hat{\mathbf{x}}_{best}$ with simulator;

05: Record the evaluation into database D;

**Output**: Database D

### 4.4 Optimization of shape parameter $\sigma$ of RBF and PNN

For a set of predefined database $\mathbf{D} = \{(\mathbf{x}_i, f(\mathbf{x}_i)), i=1,2,...,N\}$, the approximation accuracy of RBF and the classification accuracy of PNN are impacted by the shape factor value $\sigma$ of the kernel functions. Fig. 4 indicates the prediction impact of using different shape factors. To improve the approximation and classification accuracy, leave-one-out method is adopted by minimizing the validation error to determine the shape factor value. The pseudo-code of this section is shown in Algorithm 4. The validation error of the shape factor $\sigma$ is defined as:

$$e(\sigma) = \sum_{i=1}^{N} e_i = \sum_{i=1}^{N} (f(\mathbf{x}_i) - \hat{f}(\mathbf{x}_i))^2 \qquad (19)$$

where $e_i$ is the prediction error at the $i^{th}$ sample point. Therefore, the sub-optimization problem can be defined as:

$$\min_{\sigma} e(\sigma) \qquad (20)$$

subject to:

$$\sigma > 0 \quad (21)$$

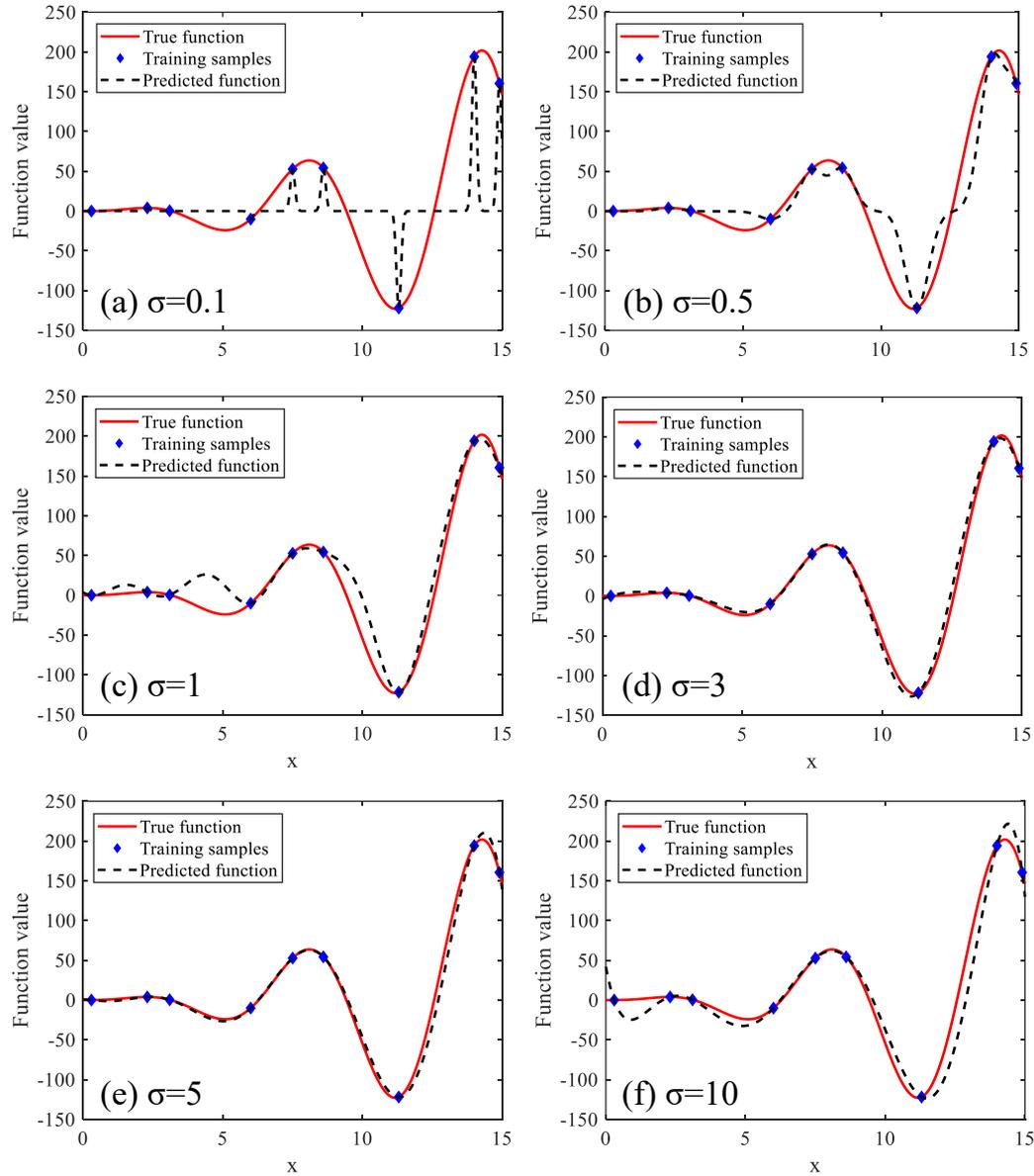

Fig. 4 The prediction impact of using different shape factors

Although the sub-optimization problem is one-dimensional, it is still computationally expensive since each function evaluation needs to train the model and calculate the prediction error at testing sample points. According to several numerical experiments conducted by Rippa [58], the $e(\sigma)$ is an approximate quadratic function. Thus, in this work, second-order PRS is adopted as surrogate model to approximate $e(\sigma)$. Specifically, sample $m$ uniformly distributed shape factors $\{\sigma_1, \sigma_2, ..., \sigma_m\}$, calculate corresponding errors $\{e(\sigma_1), e(\sigma_2), ..., e(\sigma_m)\}$ according to Eq. (19). Subsequently, the coefficient vector $\boldsymbol{\beta} = [\beta_0, \beta_1, \beta_{11}]^T$ of the PRS can be calculated by Eq. (11). Thus, the optima of the surrogate is $\sigma^* = -\dfrac{\beta_1}{2\beta_{11}}$, and the optimal shape factor is chosen as:

$$\sigma = \min\{\max\{\sigma^*, \sigma_{\min}\}, \sigma_{\max}\} \tag{22}$$

**Algorithm 4** Pseudo code of shape parameter optimization

---

**Input**: Number of training shape factors m, upper bound $\sigma_{\max}$ and lower bound $\sigma_{\min}$ of shape parameter.

01: Select m training shape factors $\{\sigma_1, \sigma_2, ..., \sigma_m\}$ uniformly from $[\sigma_{\min}, \sigma_{\max}]$;

02: Evaluate $\{e(\sigma_1), e(\sigma_2), ..., e(\sigma_m)\}$ based on Eq. (19);

03: Calculate $\boldsymbol{\beta} = [\beta_0, \beta_1, \beta_{11}]^T$ based on Eq. (11);

04: Calculate $\sigma^* = -\dfrac{\beta_1}{2\beta_{11}}$;

05: Choose the optimal shape factor $\sigma = \min\{\max\{\sigma^*, \sigma_{\min}\}, \sigma_{\max}\}$;

**Output**: Optimal shape factor $\sigma$

---

## 5 Case study

To show the effectiveness of the proposed algorithm, numerical experiments are conducted on a well-placement optimization problem of 2D channelized reservoir and a joint optimization problem of Egg model. The proposed algorithm is also compared with commonly used differential evolution (DE), GDDE with only classification-based prescreening strategy (namely Classifier) and GDDE with only local search strategy (namely Local search).

**5.1 Case 1: well-placement optimization of 2D reservoir**

A 2D channelized reservoir is chosen to test the performance of the proposed algorithm on well-placement optimization problems. The model contains $100 \times 100 \times 1$ grid blocks with the size of each grid block $100 \times 100 \times 20 \text{ft}^3$. Fig. 5 presents the permeability distribution of the 2D reservoir model. The reservoir contains several distinct high-permeability channels. There are totally 10 wells, with 5 water-injection wells and 5 production wells. Since the reservoir has no aquifer or gas cap, primary oil production is negligible. The initial reservoir pressure is 6000 psi. The porosity of the reservoir is 0.2. The initial water saturation is 0.2. The compressibility of the reservoir is $6.9 \times 10^{-5} \text{psi}^{-1}$. The viscosity of oil is 2.2 cp. The control scheme for each injection and production well is fixed. The injection rate for each injection well is set to 1000 STB/day, while the bottom-hole-pressure for each production well is set to 3000 psi. The lifetime of the project is 7200 days, and each time step is 360 days, which means there are totally 20 time steps. The main goal is to determine the locations of 5 water-injection wells and 5 production wells to further maximize the profits in the lifetime of the reservoir project. Therefore, there are totally 20 variables for the well-placement optimization problem. The oil price is set to 80 USD/STB, the cost of water injection is 5 USD/STB, and the cost of water treatment is 5 USD/STB. The discount rate is set to 0%.

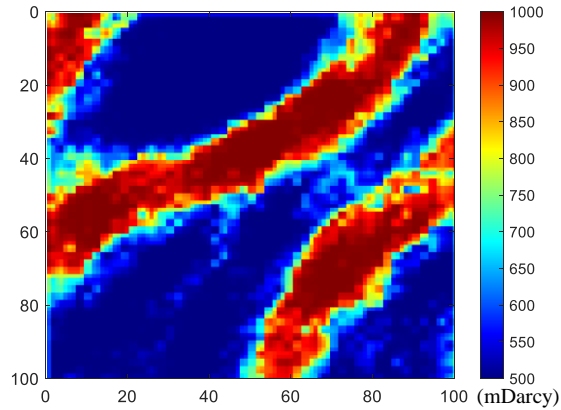

Fig. 5 The permeability distribution of the 2D reservoir

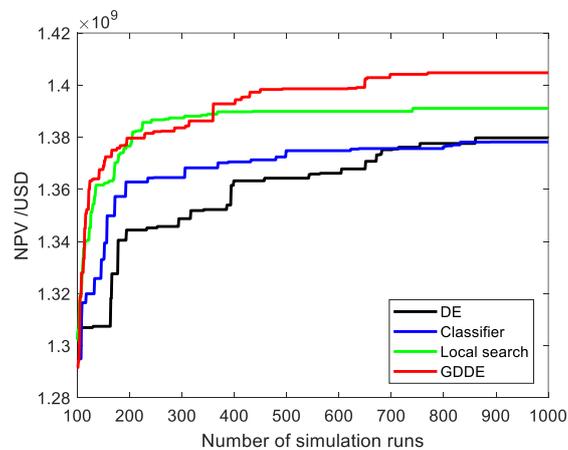

Fig. 6 NPV vs. number of simulation runs by DE, Classifier, Local search and GDDE

for case 1

To build surrogate model accurately, the initial sample points provided by LHS is set to 100 for case 1. Since the heuristic algorithms are stochastic, 5 independent runs are performed to show the performance of the algorithms. The optimization results obtained by DE, Classifier, Local search and GDDE for case 1 are shown in Fig. 6. Local search and GDDE converge fast in the early stage of optimization, while Classifier and DE converge slowly. Fig. 7 presents the distributions of convergence curves for case 1 with 5 independent runs. Classifier tends to consume more real function evaluations on exploring the uncertain area, resulting in slower convergence than DE. Local search converges efficiently, however, it is easy to get trapped into local optima. After combining classification-based prescreening strategy and local search strategy, GDDE converges efficiently and effectively. Fig. 8 presents the optimal well-placement provided by DE, Classifier, Local search and GDDE and corresponding oil saturation fields after 720, 3600 and 7200 days. As shown in Fig. 8, 3 production wells are set to the medium high-permeability channel and no injection well is set to the channel, which can reduce the water production and increase oil production. The well locations provided by GDDE can achieve more oil production and higher NPV. It is worth noting that some well locations are quite close, or even overlap (Fig. 8). Such scheme of well locations may not be optimal. If the developer wishes to avoid such phenomenon, constraints of distances between wells should be added [54].

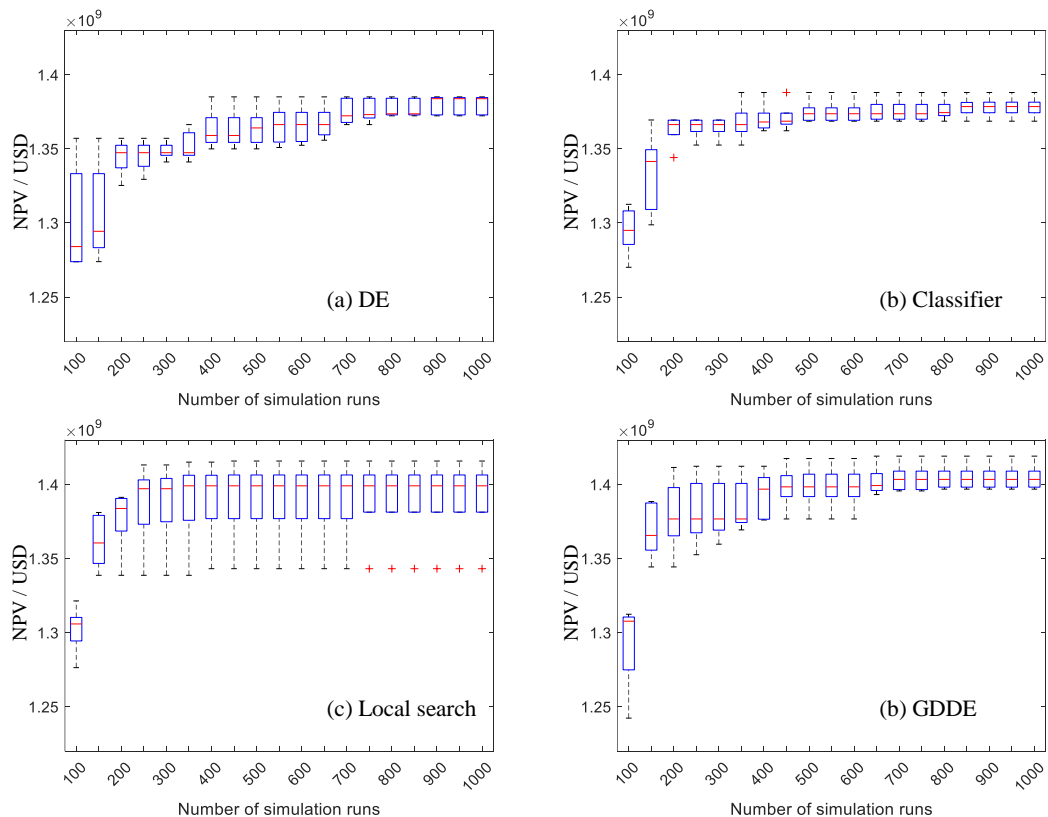

Fig. 7 The distributions of convergence curves for case 1 with 5 independent runs

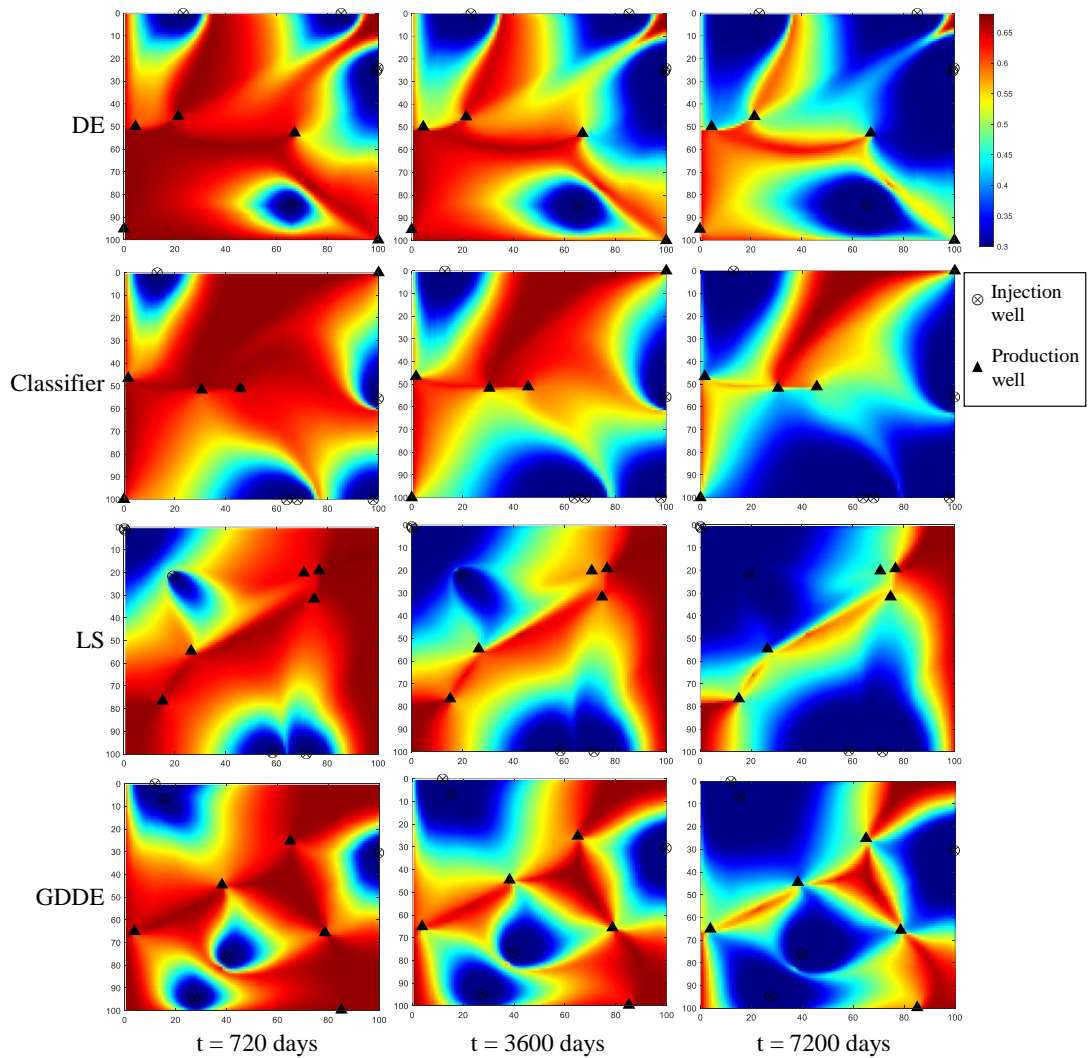

Fig. 8 The optimal well-placement by DE, Classifier, Local search and GDDE and corresponding oil saturation fields after 720, 3600 and 7200 days

**5.2 Case 2: Joint optimization of well-placement and control scheme (Egg model)**

Egg model is chosen as the 3D reservoir model for well-placement and production joint optimization problem. Egg model has been extensively used for well-placement and production optimization [7, 37, 38]. The permeability distribution of Egg model is shown in Fig. 9. The model has $60 \times 60 \times 7$ grid blocks, with the size of each grid block $30 \times 30 \times 10 \mathrm{m}^3$, and 18,853 are active grid blocks. The detailed description of Egg model can be found in Jansen et al [59]. For this case, the life cycle of the project is 3600 days, and each time step is 360 days, for a total of 10 time steps. The main goal is to determine the locations of 8 water injection wells and 4 production wells and the control scheme of 12 wells on each time step to maximize the profits in the lifetime of the Egg model. Therefore, there are totally $(8+4)\times 2+(8+4)\times 10=144$ variables for the well-placement and production joint optimization problem. The lower and upper bounds of injection wells are 0 m³/day and 80 m³/day, respectively. The lower and upper bounds of production wells are 0 m³/day and 120 m³/day, respectively. The oil price, the cost of water injection and treatment, and the discount rate are the same as case 1.

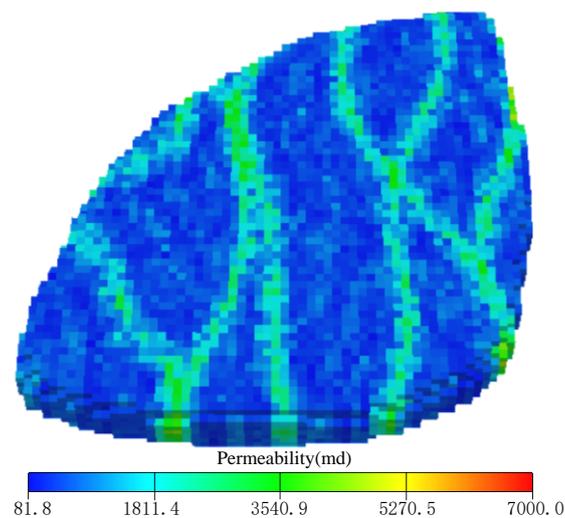

Fig. 9 The permeability distribution of Egg model

In case 2, the outline of the reservoir is not regular. When new well locations of candidate solutions generated by heuristic operators are not at active grid blocks, the well locations will be moved to the nearest active grid block. The initial sample points provided by LHS is set to 200 for the case. 5 independent runs are performed to show the statistic performance of the algorithms. The optimization results provided by DE,

Classifier, Local search and GDDE for case 2 are shown in Fig. 10. In case 2, Local search and GDDE also converge fast in comparison with Classifier and DE. Fig. 11 shows the distributions of convergence curves for case 2 with 5 independent runs. In comparison with DE, surrogate-based optimization algorithm is computationally efficient. The performance of Local search is better than Classifier, because Classifier consumes most function evaluations on exploring uncertain areas, with sparse sample points. Since it is a combinatorial optimization problem with both integer and continuous variables, the optimization problem is strongly non-linear. Therefore, the uncertainty distribution of optimization for Classifier and Local search is high (Fig. 11). While after combining classification-based prescreening strategy and local search strategy, the proposed GDDE converges efficiently, and the final result after 1000 simulation runs is also promising.

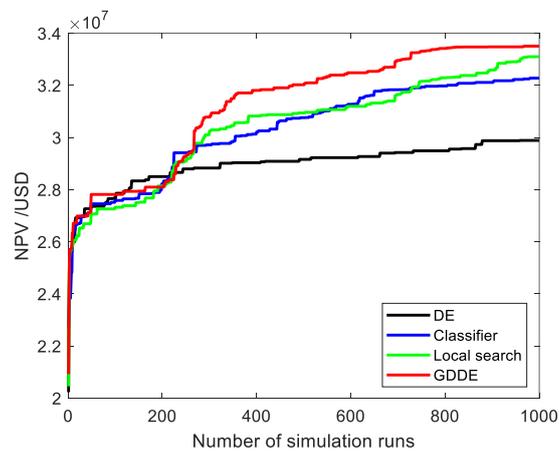

Fig. 10 NPV vs. number of simulation runs by DE, Classifier, Local search and GDDE for case 2

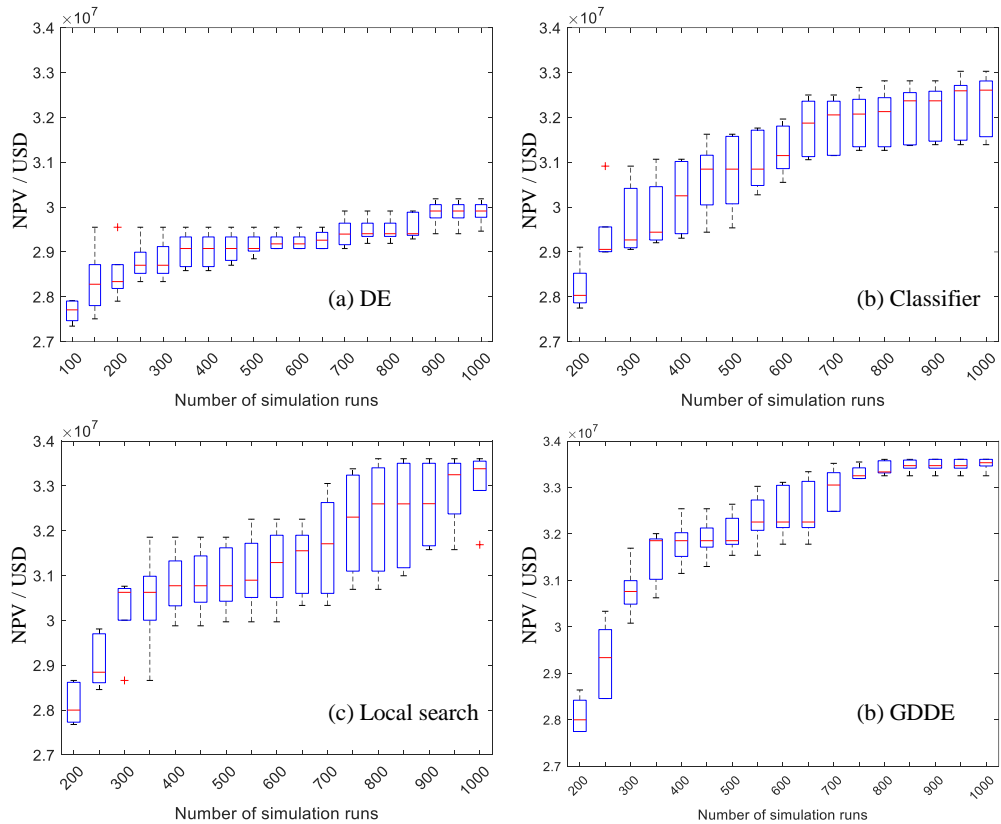

Fig. 11 The distributions of convergence curves for case 2 with 5 independent runs

Fig. 12 presents the optimal well control schemes by DE, Classifier, Local search and GDDE for Egg model. Fig. 13 indicates the simulation results of optimal control calculated by DE, Classifier, Local search and GDDE. As illustrated in Fig. 13, the developing plan of Local search can provide more oil in comparison with other methods. However, local search also leads to more cumulative water injection and treatment, which will increase the cost of reservoir development and lower the NPV. The developing plan provided by DE requires a large amount of water injection, resulting in more water output from the production wells, but no increase in cumulative oil production. The developing plan obtained by GDDE cannot significantly increase oil production compared to Local search, but can substantially reduce the water injection and water production, thereby saving development costs and improving the NPV of the project throughout its life cycle.

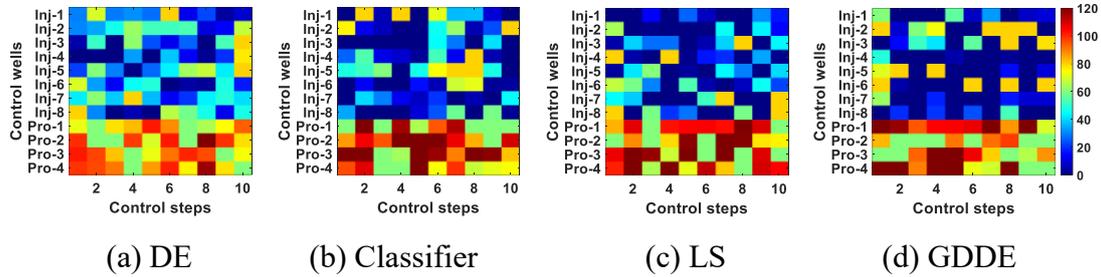

(a) DE  (b) Classifier  (c) LS  (d) GDDE

Fig. 12 The optimal well-control schemes by DE, Classifier, Local search and GDDE for Egg model

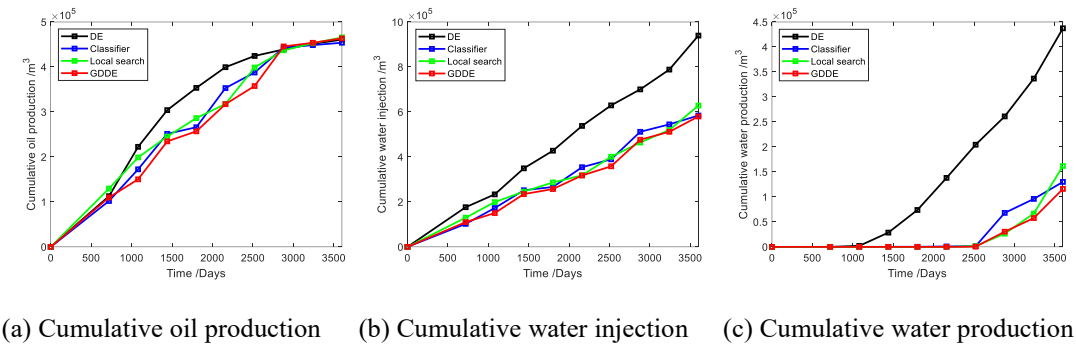

(a) Cumulative oil production  (b) Cumulative water injection  (c) Cumulative water production

Fig. 13 Simulation results of optimal control calculated by DE, Classifier, Local search and GDDE

# 6 Conclusion

In this study, a generalized data-driven differential evolutionary algorithm (GDDE) is proposed to deal with well-placement optimization or joint optimization problems. The proposed GDDE is based on classification-based and approximation-based machine learning methods combined with differential evolutionary algorithm. The proposed GDDE adopts PNN as classifier to select informative and promising candidates, and the most uncertain candidate selected by Euclidean distance is evaluated with numerical simulations. Moreover, local surrogate is adopted to accelerate the convergence. To obtain high-quality approximation model and classification model, the shape parameter of the basis function of RBF and PNN is optimized as a sub-optimization problem assisted with PRS. It is worth noting that the proposed optimization framework can incorporate other evolutionary algorithms as optimizer, such as PSO, GA, and artificial bee colony.

The proposed method is tested on the 2D reservoir well-placement optimization problem and the Egg model joint optimization problem in comparison with other heuristic algorithms and surrogate-assisted evolutionary algorithms. The performance of surrogate-assisted methods are significantly better than DE. After combining sub-expensive hyper-parameter optimization, classification-based prescreening strategy and local search strategy, the proposed GDDE converges efficiently and effectively, and the final optimization results on two cases are also promising. Therefore, the computational cost of the optimization process can be saved significantly, and the NPV of the project throughout the life cycle can be maximized. In future, the focus will be on well-placement optimization problems with no preset number of well locations and type of wells.

**Nomenclature**

$b$ = annual discount rate

$C_i$ = the i$^{th}$ class

$CR$ = crossover operator

| | | |
|---|---|---|
| $d$ | = | number of variables in vector |
| $dis$ | = | Euclidean distance between the input vectors |
| $D$ | = | Database |
| $e$ | = | validation error |
| $\hat{f}$ | = | output of surrogate model |
| $Mu$ | = | mutation operator |
| $n_1$ | = | number of variables of well location |
| $n_2$ | = | number of variables of well control |
| $n_t$ | = | time-step number |
| $n_i$ | = | number of injection wells |
| $n_p$ | = | number of production wells |
| $N_i$ | = | number of points belonging to $i^{th}$ class |
| $NP$ | = | number of individuals in the population |
| $NPV$ | = | net present value |
| $p$ | = | elapsed time, year$^{-1}$ |
| $Q_{i,t}$ | = | injection-flow-rate, STB/day |
| $Q_{o,t}$ | = | oil-production rate, STB/day |
| $Q_{w,t}$ | = | water-production rate, STB/day |
| $r_i$ | = | water-injection cost, USD/STB |
| $r_o$ | = | oil revenue, USD/STB |
| $r_w$ | = | water-removal cost, USD/STB |
| $t$ | = | time step, days |
| $\mathbf{u}$ | = | trial vector |
| $\mathbf{v}$ | = | donor vector |
| $\mathbf{x}_1$ | = | discrete integer vector of well location to be optimized |
| $\mathbf{x}_{1l}$ | = | lower boundary of well locations |
| $\mathbf{x}_{1u}$ | = | upper boundary of well locations |
| $\mathbf{x}_2$ | = | continuous vector of well control |
| $\mathbf{x}_{2l}$ | = | lower boundary of well control variables |
| $\mathbf{x}_{2u}$ | = | upper boundary of well control variables |
| $y$ | = | vector of objective function value of sample points |
| $\boldsymbol{\beta}$ | = | coefficient vector |
| $\sigma$ | = | shape factor |
| $\omega$ | = | weight parameters |
| $\varphi$ | = | basis function |
| $\tau$ | = | number of initial sample points |
| $\lambda$ | = | size of the first class |

**Superscript**

| | | |
|---|---|---|
| $j$ | = | index of element in a vector |

**Subscripts**

| | | |
|---|---|---|
| best | = | the best solution |
| $g$ | = | global surrogate |
| $i$ | = | offspring index |
| $l$ | =   | local surrogate          |
| unc | =   | the most uncertain solution |

## Acknowledgement

This study was supported by RAE Improvement Fund of the Faculty of Science, The University of Hong Kong, the grants from the Research Grants Council of Hong Kong Special Administrative Region, China, (Project No. 17303519 and 17307620). The code of the algorithm is open for access by sending Guodong Chen email.